\documentclass[11pt,a4paper]{article}
\usepackage{authblk}
\usepackage[hyperref]{acl2017}
\usepackage{times}
\usepackage{latexsym}
\usepackage{graphicx}  
\usepackage{amsmath}
\usepackage[author={Bhuwan Dhingra}]{pdfcomment}

\usepackage{url}

\aclfinalcopy 

\title{Question Answering from Unstructured Text \\ by Retrieval and Comprehension}

\author[1,2]{Yusuke Watanabe}
\author[2]{Bhuwan Dhingra}
\author[2]{Ruslan Salakhutdinov}
\affil[1]{Sony Corporation}
\affil[2]{School of Computer Science, Carnegie Mellon University}
\affil[ ]{\texttt{\{ywatanab, bdhingra, rsalakhu\}@cs.cmu.edu}}

\date{}

\begin{document}
\maketitle
\begin{abstract}
Open domain Question Answering (QA) systems must interact with external knowledge 
sources, such as web pages, to find relevant information.
Information sources like Wikipedia, however, are not well structured
and difficult to utilize in comparison with Knowledge Bases (KBs).
In this work we present a two-step approach to question answering from unstructured
text, consisting of a \textit{retrieval} step and a \textit{comprehension} step.
For comprehension, we present an RNN based attention model with a novel mixture mechanism
for selecting answers from either retrieved articles or a fixed vocabulary.
For retrieval we introduce a hand-crafted model and a neural model for ranking relevant articles.
We achieve state-of-the-art performance on \textsc{WikiMovies} dataset,
reducing the error by 40\%.
Our experimental results further demonstrate the importance of each of the introduced components.
\end{abstract}

\section{Introduction}
Natural language based consumer products, such as Apple Siri and Amazon Alexa, have found wide
spread use in the last few years. A key requirement for these conversational systems is the
ability to answer factual questions from the users, such as those about movies, music, and artists.

Most of the current approaches for Question Answering (QA) are based on structured Knowledge Bases (KB)
such as Freebase \cite{freebase:08} and Wikidata \cite{vrandevcic2014wikidata}.
In this setting the question is converted to a logical form using semantic parsing, 
which is queried against the KB to obtain the answer \cite{OQA:14,berant2013semantic}.
However, recent studies have shown that even large curated KBs, such as Freebase, are incomplete 
\cite{west2014knowledge}. Further, KBs support only certain types of answer schemas, and 
constructing and maintaining them is expensive.

\begin{figure}[t]
  \centering
  \includegraphics[bb=0 0 434 260,width=\linewidth]{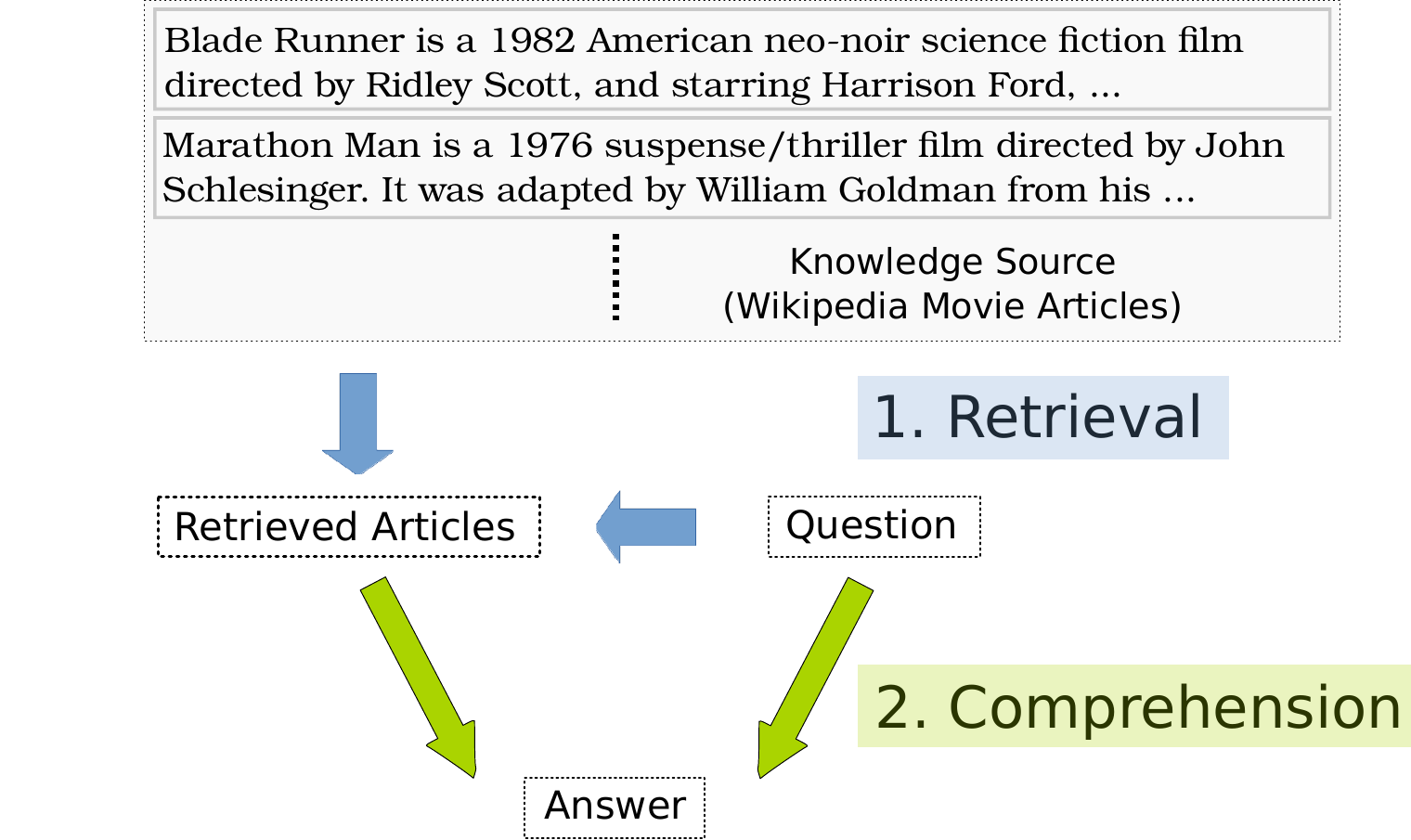}    
  \caption{\label{rc} \small  Overview of a retrieval + comprehension (r+c) QA system. First, movie articles relevant to a question are retrieved.
  Then, the retrieved articles along with the question are processed to obtain an answer.
  } 
\vspace{-0.2in}
\end{figure}

On the other hand, there is a vast amount of unstructured knowledge available in textual form
from web pages such as Wikipedia, and hence an alternative is to directly answer questions
from these documents. In this approach, shown in Figure~\ref{rc},
articles relevant to the question are first selected ({\it retrieval} step). 
Then, the retrieved articles and question are jointly processed to extract the answer
({\it comprehension} step).
This retrieval based approach has a longer history than the KB based approach \cite{voorhees:00}.
It can potentially provide a much wider coverage over questions, and is not limited to specific answer
schemas.
However,
there are still gaps in its performance compared to the KB-based approach \cite{KV:16}.
The comprehension step, which requires
parsing information from natural language, is the main bottleneck, though suboptimal
retrieval can also lead to lower performance.

\begin{figure}[t]
  \centering
  \includegraphics[bb=0 0 335 176,width=\linewidth]{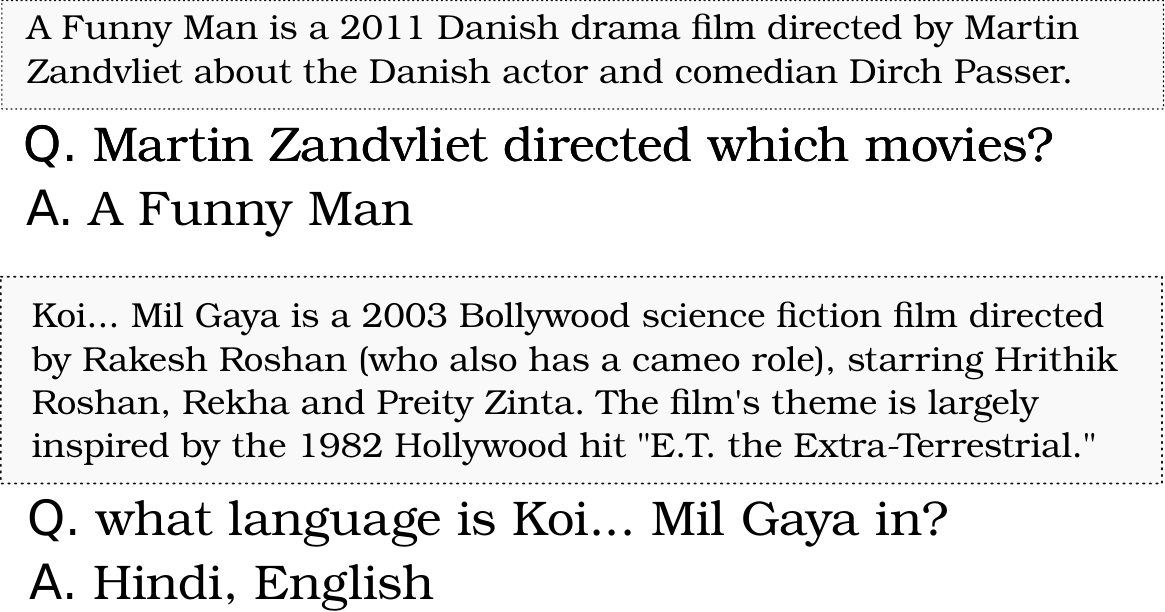}    
  \caption{\label{fig:koi_martin} \small Example of comprehension step from \textsc{WikiMovies} dataset.
  \textit{Top}: answer is a span of text in article.
  \textit{Bottom}: answer is not explicitly written in article. 
  } 
\vspace{-0.1in}
\end{figure}

Several large-scale datasets introduced recently \cite{rajpurkar2016squad,teaching:15} 
have facilitated the development of powerful neural models for reading comprehension.
These models fall into one of two categories: (1) those which extract answers as a span of text 
from the document \cite{GA:16,AS:16,xiong2016dynamic} (Figure~\ref{fig:koi_martin} top); 
(2) those which select the answer from a fixed vocabulary
\cite{chen2016thorough,KV:16} (Figure~\ref{fig:koi_martin} bottom).
Here we argue that depending on the type of question, either (1)
or (2) may be more appropriate, and introduce a latent variable mixture model to combine the two in a
single end-to-end framework.

We incorporate the above mixture model in a simple Recurrent Neural Network (RNN)
architecture with an attention mechanism \cite{bahdanau2014neural} for comprehension. 
In the second part of the paper we focus on the retrieval step for the QA system, and
introduce a neural network based ranking model to select the articles
to feed the comprehension model. 
We evaluate our model on \textsc{WikiMovies} dataset,
which consists of 200K questions about movies, along with 
18K Wikipedia articles for extracting the answers. 
\newcite{KV:16} applied Key-Value Memory Neural Networks (KV-MemNN) to the dataset,
achieving 76.2\% accuracy. 
Adding the mixture model for answer selection improves
the performance to 85.4\%.
Further, the ranking model improves both precision and recall of the retrieved articles, 
and leads to an 
overall performance of 85.8\%.

\begin{table}[t]
\begin{center}
\small
\begin{tabular}{|l|l|}
\hline 
\bf Question & \bf Answers \\ 
\hline\hline
who directed the movie Blade Runner?   &  Ridley Scott  \\
\hline
what movies can be described by & Precious, \\   
mariah carey?  & Glitter \\
\hline
what kind of film is The Hitman? & Action, Crime \\  
\hline
\end{tabular}
\end{center}
\vspace{-0.1in}
\caption{\label{table:qa} \small Example of questions and answers.}
\vspace{-0.1in}
\end{table}

\section{\textsc{WikiMovies} Dataset} \label{sec:dataset}

We focus on the \textsc{WikiMovies}\footnote{\scriptsize \url{http://fb.ai/babi}} 
dataset, proposed by \cite{KV:16}.
The dataset consists of pairs of questions and answers
about movies. 
Some examples are shown in Table~\ref{table:qa}.

As a knowledge source
approximately 18K articles from Wikipedia are also provided, where
each article is about a movie.
Since movie articles can be very long,
we only use the first paragraph of the article, which typically provides a summary of the movie.
Formally, the dataset consists of
question-answer pairs $\{(q_j, A_j)\}_{j=1}^J$ and movie articles $\{d_k\}_{k=1}^K$.
Additionally, the dataset includes a list of entities:
movie titles, actor names, genres etc.
Answers to all the questions are in the entity list.
The questions are created by human annotators
using SimpleQuestions \cite{largescale:15},
an existing open-domain question answering dataset, and
the annotated answers come from facts in two structured KBs:
OMDb\footnote{\scriptsize \url{http://beforethecode.com/projects/omdb/download.aspx}}
and 
MovieLens\footnote{\scriptsize \url{http://grouplens.org/datasets/movielens/}}.

There are two splits of the dataset. 
The ``Full'' dataset consists of 200K pairs of questions and answers.
In this dataset, some questions are difficult to answer from Wikipedia articles alone.
A second version of the dataset, ``Wiki Entity'' is constructed by 
removing those QA pairs where the entities in QAs are not found in corresponding Wikipedia articles.
We call these splits \textsc{WikiMovies-FL} and \textsc{WikiMovies-WE}, respectively.
The questions are divided into train, dev and test such that
the same question template does not appear in different splits.
Further, they can be categorized into 13 categories, including \texttt{movie\_to\_actors},
\texttt{director\_to\_movies}, etc.\footnote{Category labels are only available for dev/test dataset}
The basic statistics of the dataset are summarized in Table~\ref{movieqa}.

We also note that 
more than 50\% of the entities appear less than 5 times in the training set.  
This makes it very difficult to learn the global statistics of each entity,
necessitating the need to use an external knowledge source.

\begin{table}[t]
\begin{center}
\small
\begin{tabular}{|l|r|}
\hline
\# of questions (train/dev/test)&   \\
\ \ \ \ FL train/dev/test  &  196453/10K/10K \\
\ \ \ \ WE  train/dev/test &  96185/10K/10K \\
Avg. \# words in question &   7.7 \\   
Avg. \# of answers &  1.9 \\  

\hline
\hline

\# of movie articles &  18127  \\
Avg. \# words in article &  90.9  \\

\hline
\hline

Vocabulary     &  61696  \\   
\# of entities  &  71996  \\   
\hline
\end{tabular}
\end{center}
\vspace{-0.1in}
\caption{\label{movieqa} \small Basic statistics of \textsc{WikiMovies} dataset. }
\vspace{-0.1in}
\end{table}

\section{Comprehension Model}  \label{sec:comprehension}
Our QA system answers questions in two steps, as shown in Figure~\ref{rc}.
The first step is \textit{retrieval}, where 
articles relevant to the question are retrieved.
The second step is \textit{comprehension}, where 
the question and retrieved articles are processed to derive answers.

In this section we focus on the comprehension model, 
assuming that relevant articles have already been retrieved and merged into a \textit{context} document.
In the next section, we will discuss approaches for retrieving the articles. 

\citet{KV:16}, who introduced \textsc{WikiMovies} dataset,
used an improved variant of Memory Networks called Key-Value Memory Networks.
Instead, we use RNN based network, which has been successfully used in
many reading comprehension tasks~\cite{AS:16,GA:16,chen2016thorough}. 

\textsc{WikiMovies} dataset has two notable differences from 
many of the existing comprehension datasets, such as CNN and SQuAD \cite{AS:16,GA:16,chen2016thorough}.
First, with imperfect retrieval, the answer may not be present in the context.  
We handle this case by using the proposed mixture model.
Second, there may be multiple answers to a question, such as a list of actors.
We handle this by optimizing a sum of the cross-entropy loss over all possible answers.

We also use
attention sum architecture proposed by \citet{AS:16}, which has been shown to give high performance 
for comprehension tasks.
In this approach, attention scores over the context entities are used as the output.
We term this the attention distribution $p_{att}$, defined over the entities in the context.
The mixture model combines this distribution with another output probability 
distribution $p_{vocab}$ over all the entities in the vocabulary.
The intuition behind this is that named entities (such as actors and directors) 
can be better handled by the attention part, 
since there are few global statistics available for these, 
and other entities (such as languages and genres) can be captured by vocabulary part, 
for which global statistics can be leveraged.

\begin{figure*}[t]
  \centering
  \includegraphics[bb=0 0 479 83,width=0.7 \linewidth]{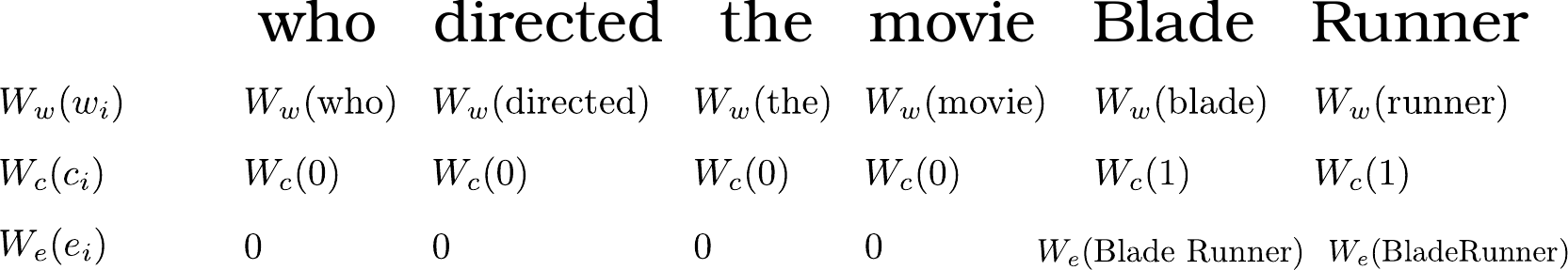}
 \vspace{-0.1in}
  \caption{\label{fig:embedding} \small Example of embedded vectors for a question ``who directed the movie Blade Runner?''}
\end{figure*}

\begin{figure*}[t]
  \centering
  \includegraphics[bb=0 0 939 152,width=0.9 \linewidth]{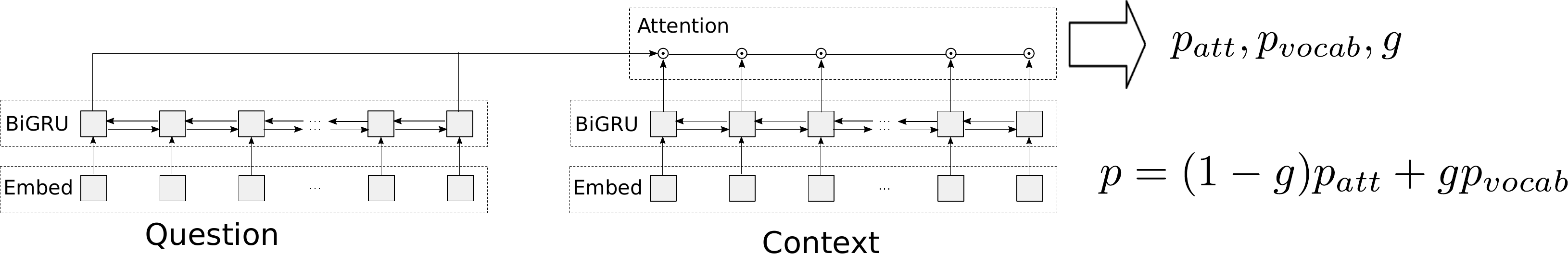}    
 \vspace{-0.1in}
  \caption{\small  \label{fig:model} Visualization of our model. 
  A question is encoded to a vector by a BiGRU. 
  With this vector, attention is computed over another BiGRU.
  Output probabilities $p_{att}, p_{vocab}$ and the mixture coefficient $g$ are
  computed from those attentions and BiGRU states.} 
 \vspace{-0.1in}
\end{figure*}

\subsection{Comprehension model detail}
Let $\mathcal{V}$ be the vocabulary consisting of all tokens in the corpus, and
$\mathcal{E}$ be the set of entities in the corpus
The question is converted to a sequence of lower cased word ids, $(w_i) \in \mathcal{V}$ and
a sequence of 0-1 flags for word capitalization, $(c_i) \in \{0,1\}$.
For each word position $i$, we also associate an entity id if the i-th word is part of
an entity, $e_i \in \mathcal{E}$ (see Figure~\ref{fig:embedding}). 
Then, the combined embedding of the i-th position is given by 
\begin{equation}
 x_i = W_w(w_i) + W_c(c_i) \Vert  W_e(e_i),  \hspace{0.1in} (i=1,\ldots,L_q),  \label{eq:embedding}
\end{equation}
where $\Vert$ is the concatenation of two vectors, 
$L_q$ is the number of words in a question $q$, and
$W_w, W_c$ and $W_e$ are embedding matrices.
Note that if there are no entities at i-th position,
$W_e(e_i)$ is set to zero.
The context is composed of up to $M$ movie articles concatenated with a special separation symbol.
The contexts are embedded in exactly the same way as questions, sharing the embedding matrices.

To avoid overfitting, we use another technique called \textit{anonymization}.
We limit the number of columns of $W_e$ to a relatively small number, $n_e$, and
entity ids are mapped to one of $n_e$ columns randomly (without collision).
The map is common for each question/context pair but randomized across pairs.
The method is similar to the anonymization method used in CNN / Daily Mail datasets~\cite{teaching:15}.
\newcite{emergent:16} showed that such a procedure actually helps readers since it adds
coreference information to the system.

Next, the question embedding sequence $(x_i)$ is fed
into a bidirectional GRU (BiGRU) \cite{cho2014learning} to obtain a fixed length vector $v$
\begin{equation}
 v = \overrightarrow{h}_{q}(L_q) \Vert  \overleftarrow{h}_{q}(0),   \label{eq:v}
\end{equation}
where $\overrightarrow{h}_{q}$ and $\overleftarrow{h}_{q}$
are the final hidden states of forward and backward GRUs respectively.

The context embedding sequence is fed into another BiGRU, to produce
the output $H_c = [h_{c,1}, h_{c,2}, \ldots h_{c,L_c}]$, where $L_c$ is
the length of the context.
An attention score for each word position~$i$ is given by 
\begin{equation}
 s_i \propto   \exp ( v^T h_{c,i} ). 
\end{equation}
The probability over the entities in the context is then given by 
\begin{equation}
 p_{att}(e) \propto   \sum_{i \in I(e, c)} s_i,
\end{equation}
where $I(e,c)$ is the set of word positions in the entity $e$ within the context $c$.

We next define the probability $p_{vocab}$ to be the probability over the complete set of entities in the corpus,
given by 
\begin{equation}
 p_{vocab}(e) = {\rm Softmax}(V u),  \label{eq:softmax}
\end{equation}
where the vector $u$ is given by 
 $u  = \sum_{i} s_i h_{c, i}$.
Each row of the matrix $V$ is the coefficient vector for an entity in the vocabulary.
It is computed similar to Eq.~(\ref{eq:embedding}).
\begin{equation}
 V(e) = \sum_{w \in e} W_w(w) + \sum_{c \in e} W_c(c) \Vert W_e(e).   \label{eq:vemb}
\end{equation}
The embedding matrices are shared between question and context.

The final probability that an entity $e$ answers the question is given by
the mixture $p(e) = (1-g) p_{att}(e) + g p_{vocab}(e)$, with the 
mixture coefficient $g$ defined as
\begin{align}
g  = \sigma(W_g g_0), \hspace{0.1in} g_0 = v^T u \Vert \max V u.
\end{align}

\noindent 
The two components of $g_0$ correspond to the attention part and
vocabulary part respectively. Depending on the strength of each, the 
value of $g$ may be high or low.

Since there may be multiple answers for a question, 
we optimize the sum of the probabilities:
\begin{equation}
 \textrm{loss} = - \log \Big( \sum_{a \in A_j} p(a|q_j,c_j) \Big)  \label{eq:loss}
\end{equation}
\noindent 
Our overall model is displayed in Figure~\ref{fig:model}.

We note that KV-MemNN~\citep{KV:16} employs ``Title encoding'' technique,
which uses the prior knowledge that movie titles are often in answers.
\citet{KV:16} showed that this technique substantially improves model performance by over 7\%
for \textsc{WikiMovies-WE} dataset.
In our work, on the other hand, we do not use any data specific feature engineering.


\begin{figure*}[t]
  \centering
  \includegraphics[bb=0 0 650 177,width=0.85\linewidth]{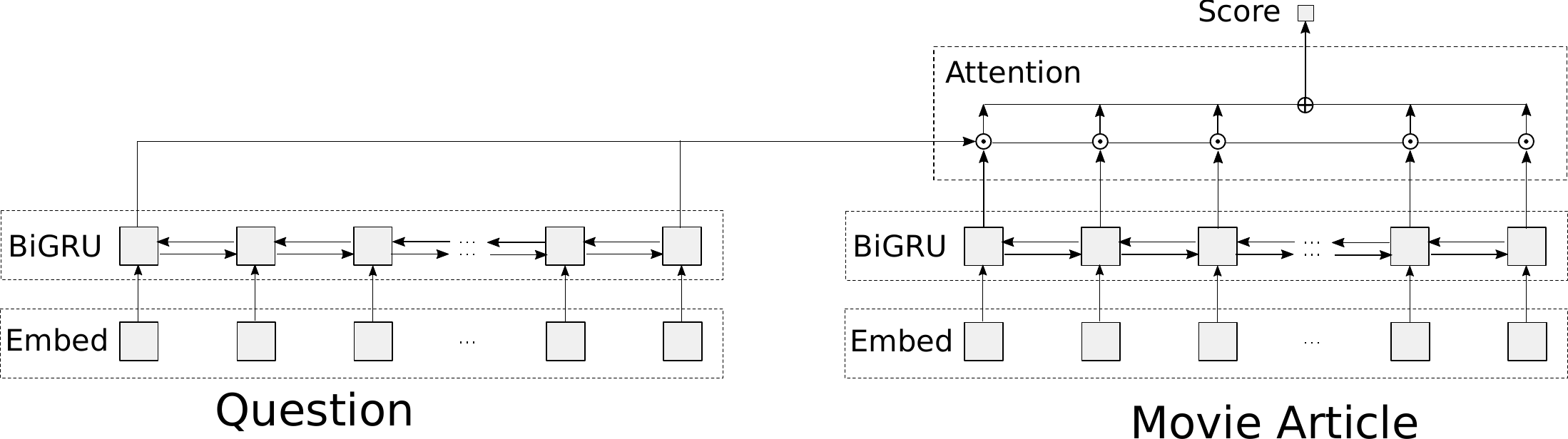}
  \vspace{-0.05in}
  \caption{\small Overview of retrieval model. 
  Similar to the comprehension model, a question is encoded to a fixed length vector.
  Attention is computed over the words of the movie article.} \label{fig:ret_model}
  \vspace{-0.1in}
\end{figure*}

\section{Retrieval Model}  \label{sec:retrieval}
Our QA system answers questions by two steps as in Figure~\ref{rc}. Accurate retrieval of 
relevant articles is essential for good performance of the comprehension model, and in this
section we discuss three approaches for it.
We use up to $M$ articles as context.
A baseline approach for retrieval is to select articles which contain at least one entity
also present in the question. 
We identify maximal intervals of words that match entities in questions and articles.
Capitalization of words is ignored in this step because some words in the questions are not properly capitalized.
Out of these (say $N$) articles we can randomly select $M$. We call this approach (r0).
For some movie titles, however, this method retrieves too many articles
that are actually not related to questions.
For example, there is a movie titled ``Love Story''
which accidentally picks up the words ``love story''.
This degrades the performance of the comprehension step.
Hence, we describe two more retrieval models -- (1) a dataset specific hand-crafted approach, and (2) a general learning based approach.

\subsection{Hand-Crafted Model (r1)}
In this approach, the $N$ articles retrieved using entity matching are assigned scores
based on certain heuristics. 
If the movie title matches an entity in the question, the article is given a high score, since 
it is very likely to be relevant. A similar heuristic was also employed in \cite{KV:16}.
In addition, the number of matching entities is also used to score each article.
The top $M$ articles based on these scores are selected for comprehension.
This hand-crafted approach already gives strong performance for the \textsc{WikiMovies} 
dataset, however the heuristic for matching article titles may not be appropriate for other 
QA tasks. Hence we also study a general learning based approach for retrieval.

\subsection{Learning Model (R2)}


The learning model for retrieval is trained by an oracle constructed using distant supervision.
Using the answer labels in the training set, we can find appropriate articles that include the information requested in the question.
For example, for \texttt{x\_to\_movie} question type, 
the answer movie articles are the correct articles to be retrieved.
On the other hand, for questions in \texttt{movie\_to\_x} type,
the movie in the question should be retrieved.
Having collected the labels, we train a retrieval model for classifying a question and article pair as relevant or not relevant.


\begin{table*}[t]
\begin{center}
\small
\begin{tabular}{|l|r|r|r|r||r|r|r|r|}
\hline \bf Model Type & \bf R@1 & \bf  R@10  & \bf R@30 & \bf R@100 & \bf P@1  & \bf P@10  & \bf P@30 & \bf P@100 \\
\hline\hline
Entity Matching Baseline (r0) &  0.733 & 0.937 & 0.963 & 0.985 & 0.642 & 0.917 & 0.958 & 0.983 \\
Entity Matching + Rule (r1) & 0.942 &  0.994 & 0.998  & 0.999 & 0.827 & 0.979  & 0.996 & 0.999 \\
Entity Matching + WLA (R2) &  0.957 & 0.997  & 0.999 & 0.999 & 0.835  & 0.986 & 0.999 & 0.999 \\
\hline
\end{tabular}
\end{center}
\vspace{-0.1in}
\caption{ \label{table:retrieval} \small Performance of retrieval methods. (WikiMovies-WE) }
\end{table*}

\begin{table*}[t]
\begin{center}
\small
\begin{tabular}{|l|r|r|r|r||r|r|r|r|}
\hline \bf Model Type & \bf R@1 & \bf  R@10  & \bf R@30 & \bf R@100 & \bf P@1  & \bf P@10  & \bf P@30 & \bf P@100 \\
\hline\hline
Word Level Attention  &  0.800 & 0.986  & 0.990 & 0.993 & 0.684  & 0.968 & 0.984 & 0.988 \\  
Sum of hidden state & 0.530 & 0.817  & 0.860  & 0.900 & 0.467 & 0.786 & 0.825  & 0.865 \\  
Query Free Attention & 0.628 & 0.833  & 0.873 & 0.909 & 0.556 & 0.798 & 0.835 & 0.873  \\  
\hline
\end{tabular}
\end{center}
\vspace{-0.1in}
\caption{\label{table:rankscore} \small Performance of scoring models }
\end{table*}

Figure~\ref{fig:ret_model} gives an overview of the model, which uses a Word Level Attention
(WLA)
mechanism.
First, the question and article are embedded into vector sequences, using the 
same method as the comprehension model.
We do not use anonymization here, 
to retain simplicity. Otherwise, the anonymization procedure would have to be repeated
several times for a potentially large collection of documents.
These vector sequences are next fed to a Bi-GRU, to produce the outputs $v$ (for the question)
and $H_c$ (for the document) similar to the previous section.

To classify the article as relevant or not, we introduce a novel attention mechanism to compute the score,
\begin{equation}
    s = \sum_{i} ((w \tilde{v} + b)^T \tilde{h}_{c,i})^4
 \end{equation}
Each term in the sum above corresponds to the match between the query representation and 
a token in the context. This is passed through a 4-th order non-linearity so that relevant tokens are emphasized more\footnote{
We use exponent $d=4$ here. Higher $d$ tend to have better performance.
Empirically, this approach works better than exponential and softmax non-linearities.}.
Next, we compute the probability that the article is relevant using a sigmoid:
\begin{equation}
 o = \sigma(w' s + b')
\end{equation}
In the above, $\tilde{x}$ is the normalized version (by L2-norm) of vector $x$,
$w, b, w', b'$ are scalar learnable parameters to control scales.


%

\begin{figure*}[t]
\vspace{-0.1in}
  \centering
  \includegraphics[bb=0 0 562 145,width=0.95\linewidth]{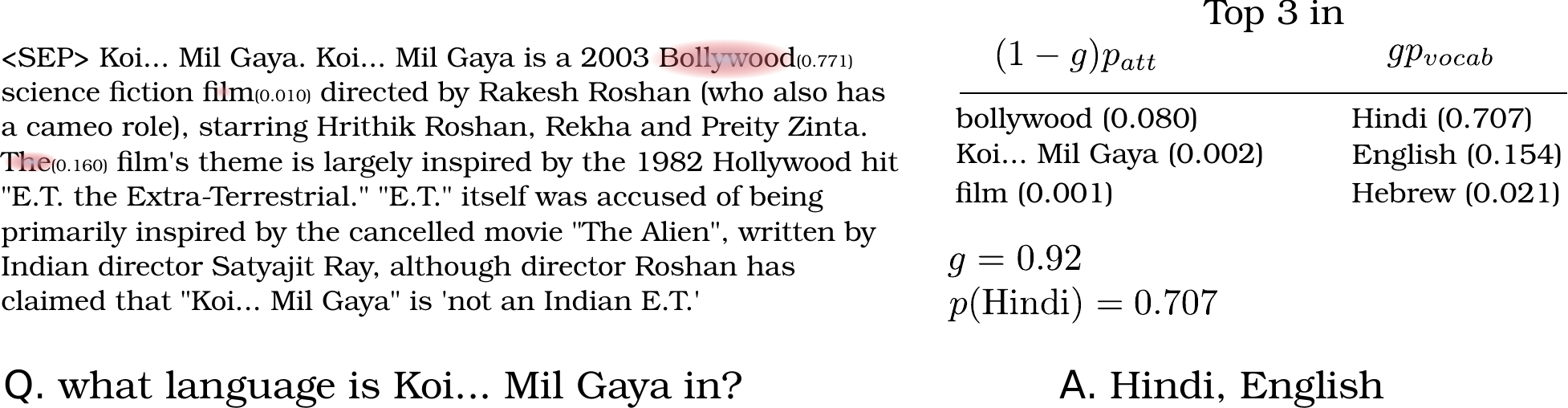}    
  \caption{ \label{fig:attention_god} \small The model uses the $p_{vocab}$ output to answer the question.
  The word ``Bollywood'' is attended. The word implies the ``Hindi'' language.
  } 
\end{figure*}

\begin{figure*}[t]
  \centering
  \includegraphics[bb=0 0 574 103,width=0.95\linewidth]{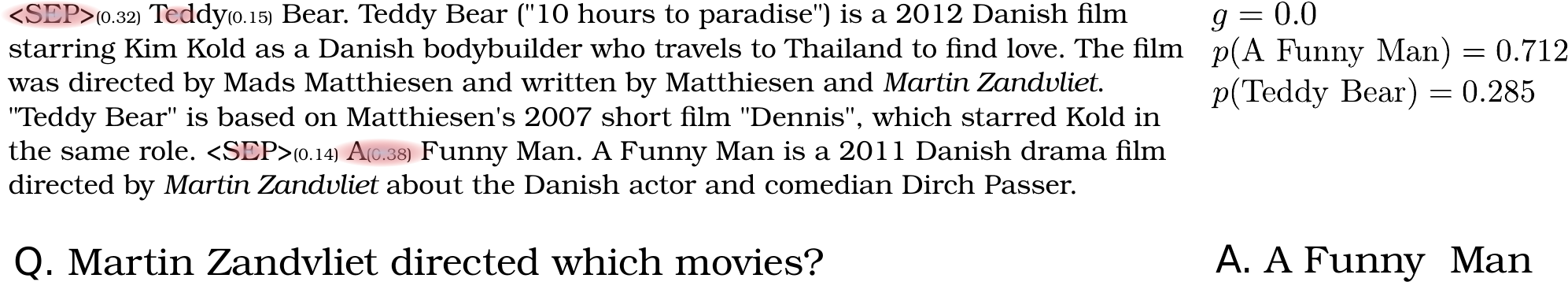}    
  \caption{\small \label{fig:attention_martin} Model behavior of a question ``Martin Zandvliet directed which movies?'' 
  Martin Zandvliet is a writer of Teddy Bear, not a director.} 
\vspace{-0.1in}
\end{figure*}

\begin{table*}[t!]
\begin{center}
\begin{tabular}{|l||r|r|r||r|r|r|}
\hline
 &\multicolumn{3}{c||}{\textsc{WikiMovies-WE}}&\multicolumn{3}{c|}{\textsc{WikiMovies-FL}}\\
\cline{2-7}
 & r0 &  r1 & R2 & r0 & r1 & R2 \\
\hline\hline
KV-MemNN &\multicolumn{3}{c||}{\textsc{ 76.2 }}&\multicolumn{3}{c|}{\textsc{ - }}\\
\hline
Vocab Model (V) & 77.5 &  81.0 & 81.9 & 54.2  & 55.8 & 57.5 \\
Attention Model (A )& 78.1 & 82.6 & 82.9 & 42.8 & 45.2 & 45.1 \\
Attention+Vocab Model (AV) & 79.4 & 83.4 & 85.1 & 58.2 & 60.4 & 60.9 \\
Attention+SubVocab Model (AsV) & 81.0 & 85.4 & 85.8 & 59.9 & 61.9 & 62.2 \\
\hline
\end{tabular}
\end{center}
\vspace{-0.1in}
\caption{\label{table:comp0} \small Performance (hits@1) comparison over different models and datasets. }
\vspace{-0.1in}
\end{table*}

\section{Experiments}  \label{sec:experiments}
We evaluate the comprehension model 
on both \textsc{WikiMovies-FL} and \textsc{WikiMovies-WE} datasets.
The performance is evaluated using the accuracy of the top hit (single answer) over all possible answers (all entities).
This is called hits@1 metric.

For the comprehension model, we use embedding dimension 100, 
and GRU dimension 128.
We use up to $M=10$ retrieved articles as context.
The order of the articles are randomly shuffled for each training instance
to prevent over-fitting. 
The size of the anonymized entity set $n_e$ is 600, since in
most of the cases, number of entities in a question and context pair is
less than 600.

For training the comprehension model, 
the Adam \citep{adam:14} optimization rule is used with batch size 32.
We stop the optimization based on dev-set performance, and
training takes around 10 epochs.
For \textsc{WikiMovies-FL} (resp. \textsc{WikiMovies-WE}) dataset, 
each epoch took approximately 4 (resp. 2) hours on an Nvidia GTX1080 GPU.

For training the retrieval model R2, we use a binary cross entropy objective.
Since most articles are not relevant to a question,
the ration of positive and negative samples is tuned to $1:10$. 
Each epoch for training the retrieval model takes about $40$ minutes
on an Nvidia GTX1080 GPU.


\subsection{Performance of Retrieval Models}
We evaluate the retrieval models based on precision and recall of the oracle articles.
The evaluation is done on the test set.
R@k is the ratio of cases where the highest ranked oracle article is in the top k retrieved articles.
P@k is the ratio of oracle articles which are in the
top k retrieved results.
These numbers are summarized in Table~\ref{table:retrieval}.
We can see that both (r1) and (R2) significantly outperform (r0), with (R2) doing 
slightly better. We emphasize that (R2) uses no domain specific knowledge, 
and can be readily applied to other datasets where articles may not be
about specific types of entities.

We have also tested simpler models based on inner product of question and article vectors.
In these models, a question $q_j$ and article $d_k$ are converted to vectors $\Phi(q_j), \Psi(d_k)$,
and the relevance score is given by their inner product:
\begin{equation}
    {\rm score}(j,k) = \Phi(q_j)^T \Psi(d_k).
 \end{equation}
In the view of computation, those models are attractive 
because we can compute the article vectors offline,
and do not need to compute the attention over words in the article.
Maximum Inner Product Search algorithms may also be utilized here
\cite{HAM:16, clusteringis:16}.
However, as shown in upper block of Table~\ref{table:rankscore},
those models perform much worse in terms of scoring.
The ``Sum of Hidden State'' and ``Query Free Attention'' models 
are similar to WLA model, using BiGRUs for question and article.
In both of those models, $\Phi(q)$ is defined the same way as WLA model, Eq~(\ref{eq:v}).
For the ``Sum of Hidden States'' model, $\Psi(d)$ is given by the sum of BiGRU hidden states.
This is the same as the proposed model by replacing the fourth order of WLA to one.
For the ``Query Free Attention'' model, $\Psi(d)$ is given by the sum of BiGRU hidden states.


We compare our model and several ablations with the KV-MemNN model.
Table~\ref{table:comp0} shows the average performance across three evaluations.
The  (V) ``Vocabulary Model'' and  (A) ``Attention Model'' are 
simplified versions of the full (AV) ``Attention and Vocabulary Model'',
using only $p_{vocab}$ and $p_{att}$, respectively.
Using a mixture of $p_{att}$ and $p_{vocab}$ gives the best performance.

Interestingly, for \textsc{WE} dataset the Attention model works better.
For \textsc{FL} dataset, on the other hand,
it is often impossible to select answer from the context, and
hence the Vocab model works better. 

The number of entities in the full vocabulary is 71K, and some of these are rare. 
Our intuition to use the Vocab model was to only use it for common entities,
and hence we next constructed a smaller vocabulary consisting of all entities which appear at least 10 times
in the corpus. 
This results in a subset vocabulary $\mathcal{V}_S$ of 2400 entities. 
Using this vocabulary in the mixture model (AsV) further improves the performance.

Table~\ref{table:comp0} also shows a comparison between (r0), (r1), and (R2) in terms
of the overall task performance.
We can see that improving the quality of retrieved articles benefits 
the downstream comprehension performance. In line with the results of the previous
section, (r1) and (R2) significantly outperform (r0). Among (r1) and (R2), (R2) 
performs slightly better.

\subsection{Benefit of training methods}
Table~\ref{table:tech} shows the impact of anonymization of entities 
and shuffling of training articles before the comprehension step,
described in Section~\ref{sec:comprehension}.

\begin{table}[t]
\begin{center}
\begin{tabular}{|l|r|r|}
\hline  
& \bf WE & \bf FL \\  
\hline\hline
r1+AsV & 85.4  &   61.9  \\
~~~~ no shuffling & 83.7  &  61.0  \\
~~~~ no anonymization & 84.5  &  61.0  \\
\hline
\end{tabular}
\end{center}
\caption{\small Shuffling and anonymization lead to higher performance.}
\label{table:tech} 
\vspace{-0.2in}
\end{table}

Shuffling the context article before concatenating them, works as a data augmentation technique.
Entity anonymization helps because without it each entity has one embedding.
Since most of the entities appear only a few times in the articles, these embeddings 
may not be properly trained.
Instead, 
the anonymous embedding vectors are
trained to distinguish different entities. This technique is motivated by a similar
procedure used in the construction of CNN / Daily Mail \citep{teaching:15}, and discussed in detail
in \citep{emergent:16}.

\subsection{Visualization}
Figure~\ref{fig:attention_god} shows a test example from the \textsc{WikiMovies-FL} test data.
In this case, even though the answers ``Hindi'' and ``English'' are not in the context,
they are correctly estimated from $p_{vocab}$. Note the high value of $g$ in this case.
Figure~\ref{fig:attention_martin} shows another example of how the mixture model works.
Here the 
the answer 
is successfully selected from the document instead of the vocabulary. Note the low value of $g$ 
in this case.

\subsection{Performance in each category}

Table~\ref{table:main} shows the comparison for each category of questions
between our model and KV-MemNN for the \textsc{WikiMovies-WE} dataset
\footnote{Categories ``Movie to IMDb Votes'' and ``Movie to IMDb Rating'' are
omitted from this table because there are only 0.5\% test data for these categories
and most of the answers are ``famous'' or ``good''.}.
We can see that performance improvements in the \texttt{movie\_to\_x} category is relatively large.
The KV-MemNN model has a dataset specific ``Title encoding'' feature
which helps the model \texttt{x\_to\_movie} question types.
However without this feature performance in other categories is poor. 

\begin{table}[t]
\begin{center}
\begin{tabular}{|l|r|r|}
\hline \bf Question Type & \bf KV & \bf  r1+AsV  \\  \hline
Movie to Year & 83 & 94  \\
Movie to Writer & 64 & 90  \\
Movie to Tags & 48 & 57  \\
Movie to Language & 84 & 89  \\
Movie to Genre & 86 & 90  \\
Movie to Director & 79 & 91  \\
Movie to Actors & 64 & 84  \\
Writer to Movie & 91 & 93  \\
Tag to Movie & 49 & 45  \\
Director to Movie & 91 & 93  \\
Actor to Movie & 83 & 85  \\
\hline
Total & \bf 76 & \bf 85.4  \\
\hline
\end{tabular}
\end{center}
\vspace{-0.1in}
\caption{\label{table:main} \small Hits@1 scores for each question type. 
Our model gets $>80\%$ in all cases but two.}
\vspace{-0.1in}
\end{table}

\subsection{Analysis of the mixture gate}

The benefit of the mixture model comes from the fact that
$p_{pointer}$ works well for some question types, 
while $p_{vocab}$ works well for others.
Table~\ref{table:g} shows how often for each category $p_{vocab}$ is used ($g > 0.5$) in AsV model.
For question types ``Movie to Language'' and ``Movie to Genre'' (the so called ``choice questions'')
the number of possible answers is small.
For this case, even if the answer can be found in the context,
it is easier for the model to select answer from an external vocabulary which encodes
global statistics about the entities.
For other ``free questions'', depending on the question type, one approach is better than the other.
Our model is able to successfully estimate the latent category 
and switch the model type by controlling the coefficient $g$.

\begin{table}[t]
\begin{center}
\begin{tabular}{|l|r||r|r|}
\hline \bf Question Type & \bf ratio & \bf r1+A & \bf r1+V \\  \hline
Movie to Year & 0.00  &  \bf 93  &  92 \\
Movie to Writer & 0.00  & \bf 90  & 86 \\
Movie to Tags &  0.01  & \bf 57  & 50  \\
Movie to Language & 0.32 & 81 & \bf 87 \\
Movie to Genre & 0.72  & 76 & \bf 90 \\
Movie to Director & 0.00 & \bf 91 & 90 \\
Movie to Actors & 0.00  & \bf 82 & 74 \\
Writer to Movie &  0.00  & \bf 92 & 89 \\
Tag to Movie & 0.03 & \bf 46  & 41 \\
Director to Movie & 0.00 & \bf 91 & 85 \\
Actor to Movie & 0.00 & \bf 81 & 80  \\
\hline
Total &   & 82.6 & 81.0 \\
\hline
\end{tabular}
\end{center}
\vspace{-0.1in}
\caption{\small Ratio of the gate being open. ($g>0.5$) ~
If the answer is named entity, the model need to select answer from text.
Therefore, $g=0$. Bold font indicates winning model. Vocabulary Only model wins when
$g$ is high.}
\label{table:g} 
\vspace{-0.1in}
\end{table}



\section{Related Work}
\newcite{hierarchical:16} solve the QA problem
by selecting a sentence in the document.
They show that joint training of selection and comprehension
slightly improves the performance.
In our case, joint training is much harder 
because of the large number of movie articles. Hence we introduce
a two-step retrieval and comprehension approach.

Recently \newcite{architecture:16} proposed a framework to
use the performance on a downstream task (e.g. comprehension) 
as a signal to guide the learning of neural network which determines the input to the 
downstream task (e.g. retrieval).
This motivates us to introduce neural network based approach for
both retrieval and comprehension, since in this case the retrieval step
can be directly trained to maximize the downstream performance.

In the context of language modeling, 
the idea of combining of two output probabilities is given in \cite{sentinel:16},
however, our equation to compute the mixture coefficient is slightly different.
More recently, \newcite{ahn2016neural}
used a mixture model to predict the next word from either the entire vocabulary, or
a set of Knowledge Base facts associated with the text. In this work, we present the first 
application of such a mixture model to reading comprehension.

\section{Conclusion and Future Work}
We have developed QA system using a two-step retrieval and comprehension approach.
The comprehension step uses a mixture model to achieve state of the art performance on \textsc{WikiMovies} dataset, 
improving previous work by a 
significant margin.

We would like to emphasize that our approach has minimal heuristics and does not use
dataset specific feature engineering.
Efficient retrieval while maintaining representation variation
is a challenging problem.
While there has been a lot of research on comprehension, 
little focus has been given to designing neural network based retrieval models. 
We present a simple such model, and emphasize the importance of this direction of research.


\bibliography{acl2017}
\bibliographystyle{acl_natbib}

\newpage
\appendix

\end{document}